\documentclass{article}
\usepackage{spconf,amsmath,graphicx}
\usepackage{seqsplit}
\usepackage{caption}
\usepackage[inline]{enumitem} 
\usepackage{mathtools}
\usepackage{amsfonts}
\usepackage{algorithm}
\usepackage{algorithmic}
\usepackage{booktabs}
\usepackage{subcaption}
\usepackage{url}

\setlist[itemize]{topsep=1pt}

\DeclareMathOperator*{\argmin}{arg\,min}
\DeclareMathOperator*{\argmax}{arg\,max}
\newcommand\numberthis{\addtocounter{equation}{1}\tag{\theequation}}
\title{Semi-supervised Batch Active Learning via Bilevel Optimization}
\name{Zal\'an Borsos$^{ \dagger}$ \qquad Marco Tagliasacchi$^{\star}$ \qquad Andreas Krause$^{\dagger}$}
\address{$^{\dagger}$ Department of Computer Science, ETH Zurich \\
      $^{\star}$ Google Research \\
      zborsos@ethz.ch \quad mtagliasacchi@google.com \quad krausea@ethz.ch}
\begin{document}
%
\maketitle
\begin{abstract}
\emph{Active learning} is an effective technique for  reducing the labeling cost by improving data efficiency. In this work, we propose a novel \emph{batch acquisition strategy} for active learning in the setting where the model training is performed in a \emph{semi-supervised} manner. We formulate our approach as a \emph{data summarization} problem via \emph{bilevel optimization}, where the queried batch consists of the points that best summarize the unlabeled data pool. We show that our method is highly effective in \emph{keyword detection} tasks in the regime when only \emph{few labeled samples} are available.
\end{abstract}
\section{Introduction}
\label{sec:intro}

Many practical applications of supervised learning face the challenge of high labeling costs  due to the involvement of human expertise. At the same time, gathering unlabeled data is often less expensive. \emph{Active learning} is an extensively studied technique for improving data efficiency, which proceeds in rounds of label acquisition and model retraining. In each round of \emph{pool-based} active learning, the goal is to select samples from the unlabeled data pool to be labeled by an expert, such that the generalization error of the model is maximally reduced when the newly acquired labels are also considered.

\looseness -1 Prominent approaches to active learning include uncertainty sampling~\cite{lewis1994sequential},  margin-based selection~\cite{balcan2007margin} and expected informativeness~\cite{mackay1992information}. 
Since acquiring labels one-by-one and retraining the model after each acquisition can be resource-intensive, batch active learning approaches~\cite{hoi2006batch, guo2008discriminative, NIPS2019_8925}  query the labels of multiple points in a single round. The challenge in this setup is to ensure the informativeness of individual points while also avoiding the redundancy in the selection.

\looseness -1 While most active learning approaches work in the pool-based setup, they consider training the model on the labeled set  and use the unlabeled data pool in the acquisition step only. Recent progress in semi-supervised learning (SSL)~\cite{rasmus2015semi, miyato2018virtual, berthelot2019mixmatch} has significantly reduced the number of labels required for training highly accurate models in the image domain.  For example, MixMatch~\cite{berthelot2019mixmatch} allows reaching $89\%$ test accuracy on CIFAR-10 with only 250 labeled points --- meanwhile training in a supervised manner on 250 points would result in under $40\%$ test accuracy. This suggests that ignoring the pool of unlabeled data and training in a supervised manner during active learning might lead to suboptimal acquisitions.

\looseness -1 The idea of combining pool-based active learning with SSL, although quite natural, has received relatively little attention. Early attempts showed improved label efficiency for Gaussian fields~\cite{DBLP:conf/icml/ZhuGL03} and SVMs~\cite{1467457, leng2013combining}. More closely related to our work, in the context of deep learning, 
 Sener et al.~\cite{sener2018active} propose to acquire labels for the points solving the $k$-center problem in the last layer embedding of the network trained in a semi-supervised manner. Song et al.~\cite{song2019combining} show that combining MixMatch with well-known acquisition functions improves label efficiency in batch active learning. Gao et al.~\cite{gao2019consistency} propose a consistency-based batch selection and show the benefit of the strategy when applied with MixMatch. We provide an empirical comparison to these methods in Section \ref{sec:experiments}.

In this work, we propose a novel \emph{batch data acquisition strategy}  via \emph{bilevel optimization} for pool-based active learning when the model training is performed in a semi-supervised manner. Inspired by the recent work of Borsos et  al.~\cite{borsos2020coresets} on data summarization, we formulate the batch acquisition as a bilevel optimization problem with cardinality constraints. In this formulation, the points selected for labeling are the ones that best summarize the ``pseudo-labeled'' data pool, i.e.,  the labels are guessed by the model trained with SSL. Similarly to~\cite{borsos2020coresets}, we approach the resulting optimization problem via forward greedy selection through a proxy reformulation, which we extend to handle data augmentations efficiently. Our formulation naturally supports batch selection and ensures diversity within the selected batch. Our main contributions are the following:
\begin{itemize}\setlength\itemsep{-0.3em}
    \item We demonstrate the effectiveness of SSL using MixMatch~\cite{berthelot2019mixmatch} in keyword detection tasks.
    \item We propose a novel data acquisition strategy for semi-supervised batch active learning.
    \item We show that our approach significantly outperforms other selection strategies in keyword detection tasks, requiring up to $30\%$ fewer labels for achieving the same performance.
\end{itemize}

\section{Method}
\looseness -1 Consider one round of batch active learning: given the labeled training set $\mathcal{D}_{\textup{train}}=\{ (x_i, y_i)\}_{i=1}^{n_{\textup{labeled}}}$ and the unlabeled data-pool $\mathcal{D}_{\textup{pool}} = \{ (x_i) \}_{i=1}^{n_{\textup{unlabeled}}}$, the  goal is to select and query the labels for a batch $\mathcal{B} \subset \mathcal{D}_{\textup{pool}}$ such that the generalization error of the learner is maximally reduced with the newly acquired labels.
Although several batch acquisition strategies have been proposed, as discussed in the previous section, the vast majority of these approaches operate in the supervised setting: the learner considers $\mathcal{D}_{\textup{train}}$ only while $\mathcal{D}_{\textup{pool}}$ is used for acquisition and is ignored during training. 

We propose a batch acquisition strategy that takes full advantage of the unlabelled data by operating in a semi-supervised setup. Oblivious to the specific algorithm used for SSL, our method relies on a single assumption: training with the SSL algorithm has lower generalization error than training in supervised manner only (Assumption 1).
While in our experiments we rely on MixMatch, we abstract the details of the SSL algorithm for the sake of the presentation.

\looseness -1 \textbf{Bilevel formulation.} Let $f$ denote the base model and $\theta_{SSL}^*$ its optimal parameters learned in a semi-supervised manner, and assume that the supervised cost function for the $c$-class classification problem is the cross-entropy loss,  denoted by $\ell$. Using the model $f_{\theta_{SSL}^*}$, we can generate soft pseudo-labels for the unlabeled pool by $\tilde{y}_x:=f_{\theta_{SSL}^*}(x)$ for all $x\in \mathcal{D}_{\textup{pool}}$. Given the labeled $\mathcal{D}_{\textup{train}}$ and the pseudo-labeled $\mathcal{D}_{\textup{pool}}$, we formulate our batch selection strategy as follows: summarize $\mathcal{D}_{\textup{pool}}$ by selecting a batch $B \subset \mathcal{D}_{\textup{pool}}$ of size $b$ such that when $f$ is trained in a \emph{supervised} manner on $\mathcal{D}_{\textup{train}} \cup B$, it generalizes well to $\mathcal{D}_{\textup{train}} \cup \mathcal{D}_{\textup{pool}}$. Formally, we select $B$ as the solution of the following optimization problem, 
\begin{equation}\label{eq:bilevel-general}
            \begin{aligned}
            &\min_{B \subset \mathcal{D}_{\textup{pool}}, |B|=b} \quad  \smashoperator{\sum_{(x,y)\in\mathcal{D}_{\textup{train}} }} \; \ell(f_{\theta^*}(x), y) +  \smashoperator{\sum_{x\in\mathcal{D}_{\textup{pool}}}}\; \ell(f_{\theta^*}(x), \tilde{y}_x)\\
            &\textrm{s.t.} \,  \theta^* \in \argmin_{\theta} \, \smashoperator{\sum_{(x,y)\in\mathcal{D}_{\textup{train}} }} \; \ell(f_{\theta}(x), y) + \smashoperator{\sum_{x\in B}}\; \ell(f_{\theta}(x), \tilde{y}_x),
            \end{aligned}
        \end{equation}
which is an instance of a cardinality-constrained bilevel optimization problem: while the lower level objective captures training in supervised manner on $\mathcal{D}_{\textup{train}} \cup B$, the upper level problem serves as a proxy for the generalization error due to Assumption 1. Let us denote the lower level objective by $F(B, \theta)$ and the upper level objective by $G(\theta)$.

The summary $B$, containing the most important points of $\mathcal{D}_{\textup{pool}}$ for supervised training, is also known as  ``coreset'' in the literature~\cite{agarwal2005geometric, bachem2017practical}. In order to motivate acquiring labels for points in $B$, consider the following cases for $x \in B$: \begin{enumerate*}[label=(\roman*)]
  \item if $f_{\theta_{SSL}^*}$ misclassifies $x$, then acquiring the correct label will induce a large model change, as $x$ belongs to the pool's most influential points;
  \item even if $f_{\theta_{SSL}^*}$ classifies $x$ correctly, it might do so with low confidence, thus acquiring hard labels for $x$ can benefit SSL method to propagate labels in the neighborhood of $x$. 
\end{enumerate*} 
One of the core challenges in batch active learning is to ensure that a diverse batch containing no redundant points is selected for labeling. We note that, under Assumption 1, this is automatically guaranteed by our formulation in Eq.~\eqref{eq:bilevel-general}, since the batch is selected to minimize a proxy (upper level objective) to the generalization error.

The combinatorial optimization problem in Eq.~\eqref{eq:bilevel-general} is an instance of the coreset generation framework recently proposed by Borsos et al.~\cite{borsos2020coresets} restricted to unweighted points. The authors propose a forward greedy heuristic based on minimizing the first order Taylor expansion of the global objective in Eq.~\eqref{eq:bilevel-general}: suppose the set of points $B'\subset \mathcal{D}_{\textup{pool}}$ has already been selected; first, the inner optimization problem $\theta_{B'}^* \in \argmin_\theta F(B', \theta)$ is solved, and the next point to be added is greedily chosen by:
\begin{equation} \label{eq:selection-rule}
    x^* {=}\!\!\argmax_{x\in \mathcal{D}_{\textup{pool}} \setminus B'}\!\! \nabla_\theta \ell(f_{\theta}(x), \tilde{y}_x)^\top\! \Bigg(\frac{\partial^2F(B', \theta)}{\partial \theta \partial \theta^\top} \Bigg)^{-1}\!\!\!\!\!\!  \nabla_\theta G(\theta),
\end{equation}
where the gradients and partial derivatives are evaluated at $\theta_{B'}^*$. Then $x^*$ is added to $B'$ and the iteration resumes with re-solving the inner optimization problem.

\looseness -1 \textbf{Proxy reformulation.} Since  in the applications of interest  $f$ is a deep neural network, the inverse-Hessian vector product in Eq.~\eqref{eq:selection-rule} is computationally intensive. The authors in~\cite{borsos2020coresets} empirically show that, for several settings, the coreset selection can be solved in reformulation via a proxy model that is related to $f$. For neural networks, the chosen proxy model relies on the corresponding Neural Tangent Kernel (NTK)~\cite{jacot2018neural}, which is a fixed kernel related to the training of the network in the infinite-width limit with gradient descent. Their reformulation, however, is only practical for small coreset sizes, as the time complexity depends cubically on the number of selected points. Moreover the reformulation does not support data augmentations,  which are crucial in the inner objective of Eq.~\eqref{eq:bilevel-general} for the good performance in our setting.

\begin{figure*}[t!]
\centering
\begin{minipage}[c]{0.59\linewidth}
\centering
\begin{subfigure}[c]{0.495\linewidth}
\centering
  \includegraphics[width=\linewidth]{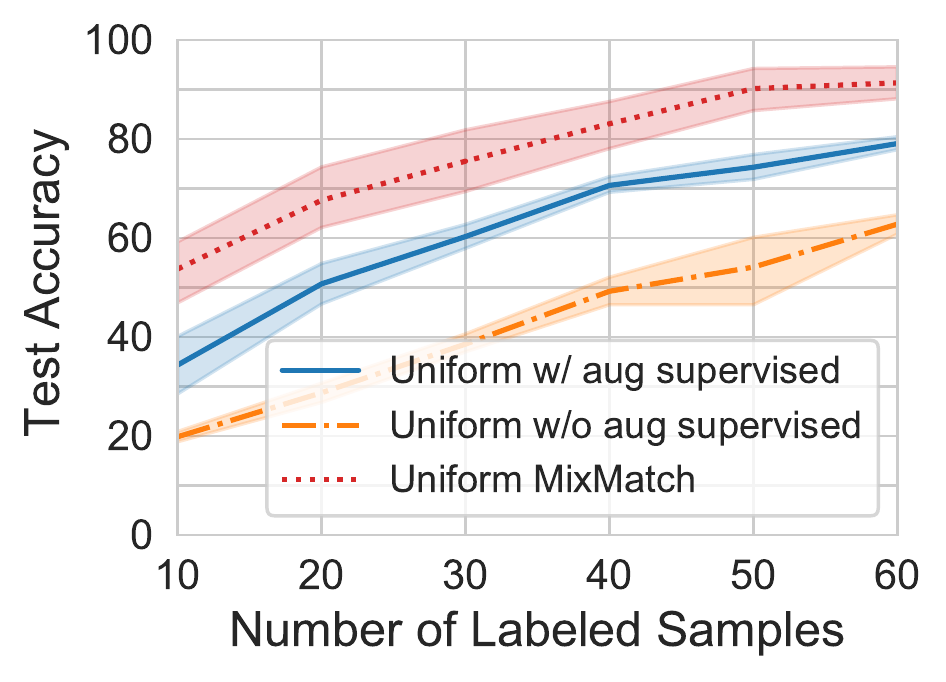} 
  \caption{Free Spoken Digit Dataset~\cite{spokendigit}}
  \label{fig:fsdd-uniform}
\end{subfigure}
\begin{subfigure}[c]{0.495\linewidth}
\includegraphics[width=\linewidth]{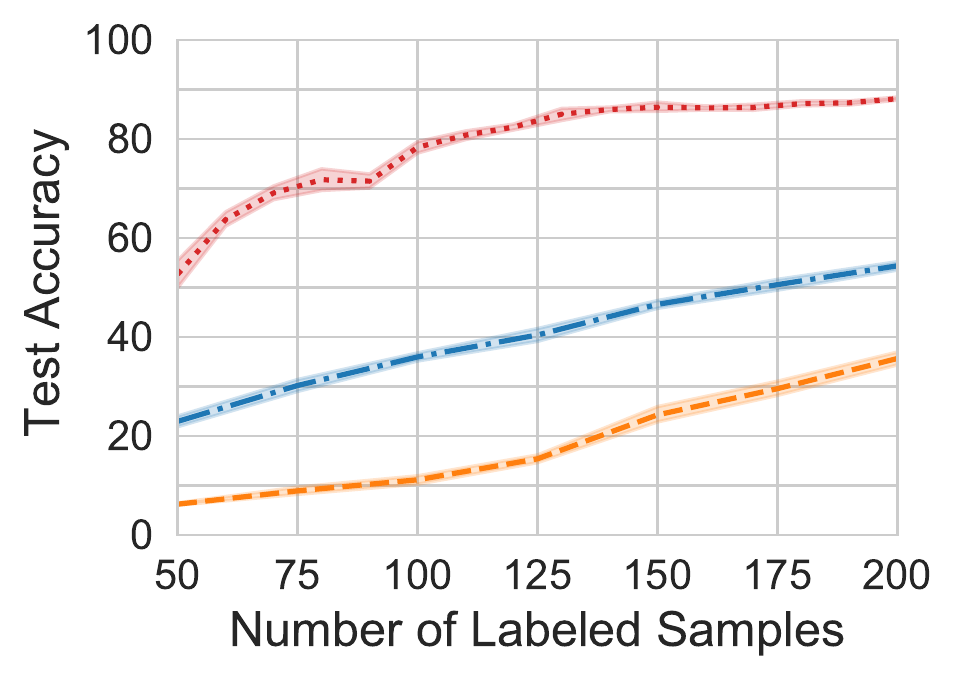} 
  \caption{Speech Commands~\cite{speechcommandsv2}}
  \label{fig:speech-commands-uniform}
  \end{subfigure}
  \caption{Supervised (with and without data augmentations) and semi-supervised learning with MixMatch~\cite{berthelot2019mixmatch} with labeled samples chosen uniformly at random.}
 \end{minipage}
  \hfill
\begin{minipage}[c]{0.40\linewidth}
\centering
  \includegraphics[width=0.9\linewidth]{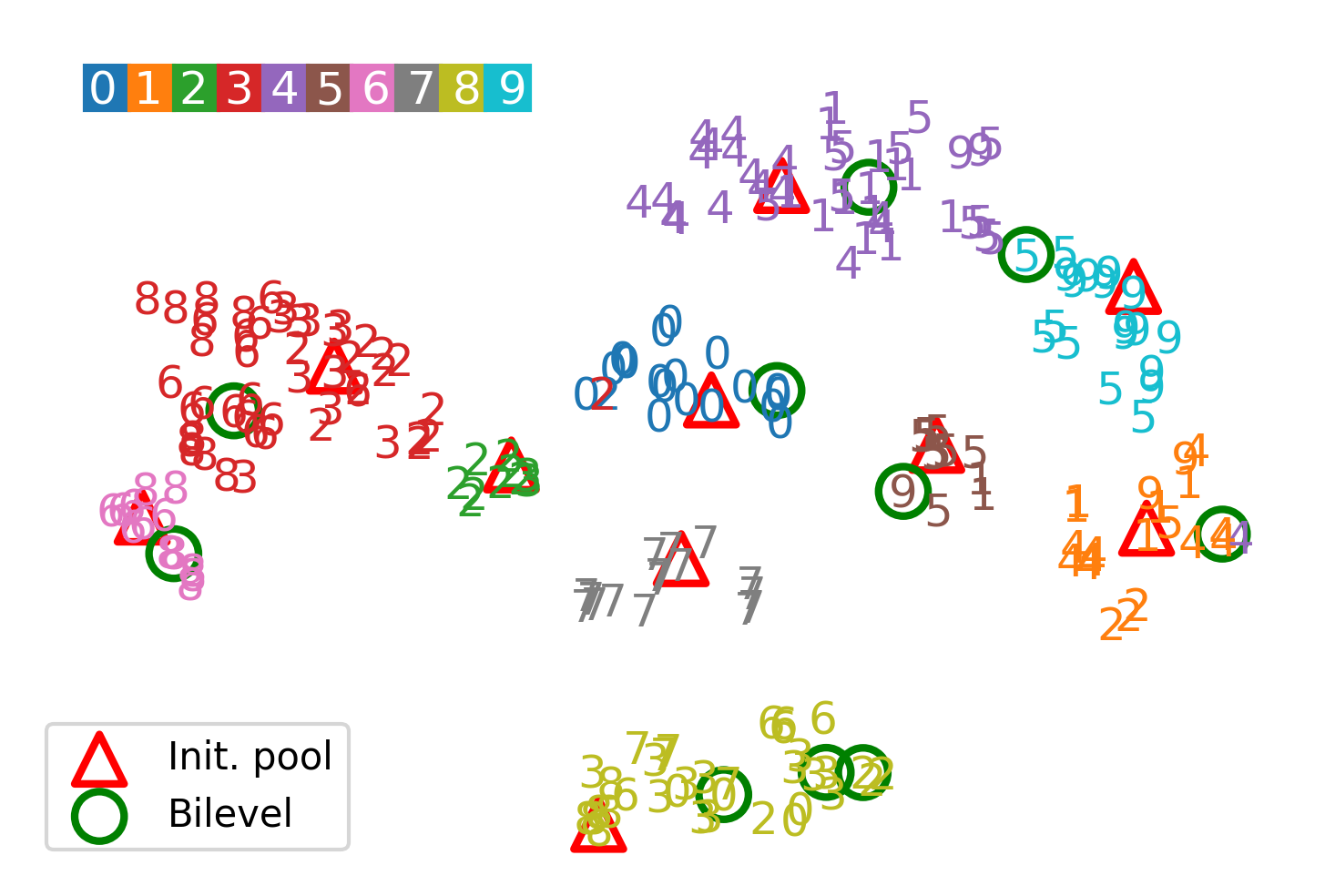}
  \caption{\looseness -1 First round of batch selection by our method (Bilevel) on Free Spoken Digit. Color codes (top left) denote predictions by the model trained on the initial pool, digits denote the true classes.}
  \label{fig:bilevel-diversity}
  \vspace{-5mm}
\end{minipage}
\end{figure*}

\looseness -1 We thus propose a proxy formulation which eliminates the cubic dependence on the number of selected points and supports data augmentations. Similarly to~\cite{borsos2020coresets}, we rely on the NTK corresponding to the neural network at hand. However, instead of using the representer theorem as in~\cite{borsos2020coresets}, we propose to low-rank approximate the kernel matrix via the Nystr\"om method. 
For mapping to Nystr\"om features, we select the subset of samples   $U=\{u_1,\dots, u_{m}\}$ from $\mathcal{D}_{\textup{train}} \cup \mathcal{D}_{\textup{pool}} $ at random, and calculate $K^U$, where $K^U_{i,j} = k(u_i, u_j)$ for $i, j \in [m]$ and $k$ is the NTK. We obtain the Nystr\"om features for $x$ by $z_x:=(K^U)^{\dagger / 2}[k(x, u_1),\dots, k(x, u_{m}) ]^\top$. 
Let us further denote $h_w(x):=\sigma(w^\top z_x)$, where  $w\in \mathbb{R}^{m\times c}$ and $\sigma$ is the softmax function. We propose the following reformulation:
\begin{align*}
&\min_{B \subset \mathcal{D}_{\textup{pool}}, |B|=b} \quad   \smashoperator{\sum_{(x,y)\in\mathcal{D}_{\textup{train}} }} \; \ell(h_{w^*}(x), y) +  \smashoperator{\sum_{x\in\mathcal{D}_{\textup{pool}}}}\; \ell(h_{w^*}(x), \tilde{y}_x) \numberthis \label{eq:proxy}\\
&\hspace*{-1mm}\textrm{s.t. }   w^* \!{=}\! \argmin_{w} \, \smashoperator{\sum_{(x,y)\in\mathcal{D}_{\textup{train}} }} \ell(h_w(x), y) \! +\! \smashoperator{\sum_{x\in B}} \ell(h_w(x), \tilde{y}_x) \!+\! \lambda \lVert w\rVert^2
\end{align*}
The inner optimization problem has thus a strongly convex objective --- multi-class logistic regression with weight decay --- related to the original neural network via working on the Nystr\"om features of the corresponding NTK. In this formulation, using data augmentations in the inner objective is straightforward. We optimize the inner objective for $nr\_it=10^3$ steps using Adam~\cite{kingma2015adam} with batch size 64 and set $\lambda=10^{-4}$. Similarly to~\cite{borsos2020coresets}, we approximate the inverse Hessian-vector product in the selection rule (Eq.~\eqref{eq:selection-rule}) via 30 conjugate gradient steps~\cite{pedregosa2016hyperparameter}. We summarize our batch selection strategy in Algorithm \ref{alg:bilevel_batch_al}. We note that we use the proxy formulation \emph{only for selecting the batch}, while we train the original model in each active learning round.
  \vspace{-2mm}
\begin{algorithm}[H]
    \caption{Batch Active Learning via Bilevel Optimization}
   \label{alg:bilevel_batch_al}
\begin{algorithmic}
   \STATE {\bfseries Input:} Labeled data $\mathcal{D}_{\textup{train}}$, unlabeled pool $\mathcal{D}_{\textup{pool}}$, model $f_{\theta_{SSL}^*}$ trained with SSL, batch size $b$, $\lambda$, $nr\_it$.
   \STATE {\bfseries Output:} Batch $B$ for label query.
   \\\hrulefill
   \STATE Generate pseudo-labels $\tilde{y}_x = f_{\theta_{SSL}^*}(x)$ for all $x\in \mathcal{D}_{\textup{pool}}$.
   \STATE Initialize $w\in \mathbb{R}^{m\times c}$ randomly, set $B=\emptyset$.
   \FOR{$\tilde{b} \in [1, ..., b]$}
    \FOR{$it \in [1, ..., nr\_it]$}
        \STATE Sample minibatch $S$ from $\mathcal{D}_{\textup{train}} \cup B$ w/ data augm.
         \STATE Generate Nystr\"om  features $\tilde{z}_x$ for all $x\in S$.
        \STATE Update $w$ by SGD on the inner obj. of Eq.\eqref{eq:proxy} with $\tilde{z}_S$.
    \ENDFOR
    \STATE Select $x^*$ by Eq.\eqref{eq:selection-rule} with $\theta$ replaced by $w$ and $f_\theta$ by $h_w$.
    \STATE Set $B = B\cup \{x^*\}$.
   \ENDFOR
\end{algorithmic}
\end{algorithm}
 \vspace{-4mm}

\begin{figure*}[ht]
\centering
\begin{subfigure}[c]{0.47\linewidth}
\centering
\includegraphics[width=0.9\linewidth]{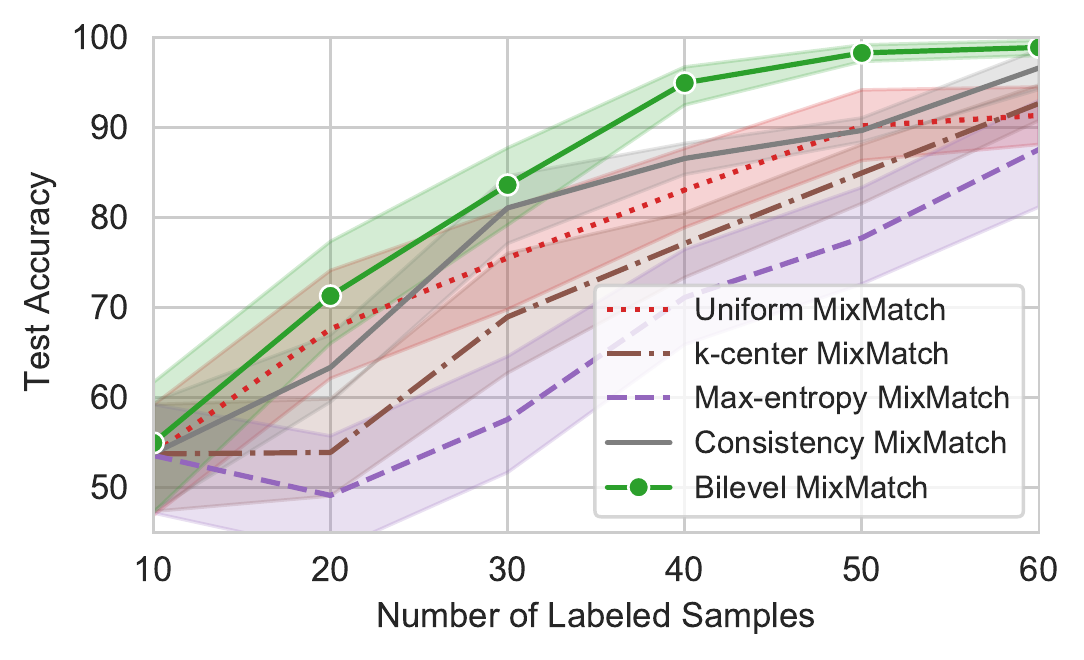}
\vspace{-2mm}
\caption{Free Spoken Digit Dataset~\cite{spokendigit}}
  \label{fig:fsdd}
\end{subfigure} \hfill
\begin{subfigure}[c]{0.47\linewidth}
\centering
\includegraphics[width=0.9\linewidth]{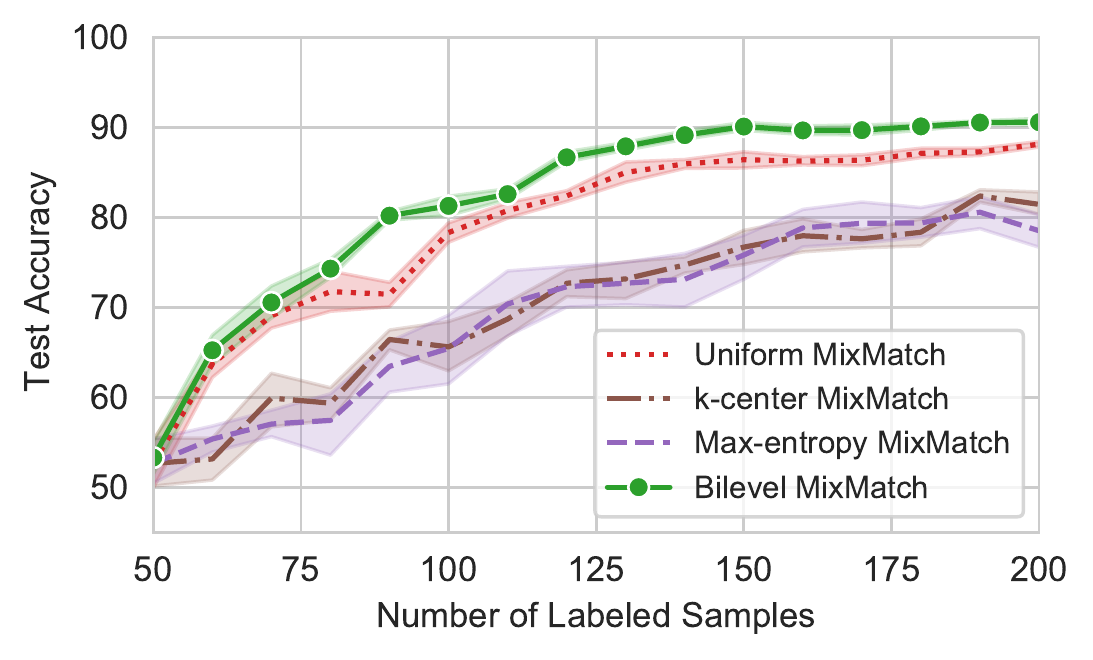}
\vspace{-2mm}
 \caption{Speech Commands~\cite{speechcommandsv2}}
  \label{fig:speech-commands}
  \end{subfigure}
  \vspace{-2mm}
  \caption{Semi-supervised batch active learning with 10 labels acquired per active learning round. We report the average test accuracy over 6 random seeds, where bands represent one standard deviation. 
  When combined with MixMatch, our proposed acquisition strategy  consistently outperforms competing approaches.}
  \vspace{-4mm}
\end{figure*}

\section{Experiments}
\label{sec:experiments}

\looseness -1 We evaluate our proposed method for keyword detection tasks on the Free Spoken Digit Dataset~\cite{spokendigit} (2700 utterances of length max 1 second, 10 classes) and on Speech Commands V2~\cite{speechcommandsv2} ($\sim$85000 utterances of 1 second, 35 classes). We pad the utterances to the length of 1 second, resample to 16 kHz and compute the mel spectrogram features with a window length of 2048 samples (128 ms), a hop length of 512 (32 ms) and with 32 bins. The resulting $32\times 32$ spectrograms allow us to use ResNets proposed in the image domain for CIFAR-10 without architectural modifications.
We employ the following augmentations independently with probability 0.5: 
\begin{enumerate*}[label=(\roman*)]
  \item random change of amplitude by a factor of $u\sim\mathcal{U}[0.8, 1.2]$,
  \item changing the speed of the audio by a factor of $u\sim\mathcal{U}[0.8, 1.2]$,
  \item random shifts in the time domain by $t\sim\mathcal{U}[-250, 250]$ ms,
  \item adding background noise sequences provided with Speech Commands with SNR $r\sim\mathcal{U}[0, 40]$ dB.
\end{enumerate*} 
We train a Wide ResNet-28~\cite{zagoruyko2016wide} without dropout on the resulting mel spectrograms.

First, we demonstrate the effectiveness of SSL in the domain of keyword detection. Our SSL algorithm of choice is MixMatch~\cite{berthelot2019mixmatch} with Wide ResNet-28, which has been show to provide large performance gains in the image domain. We use two augmentations for label guessing for MixMatch and set the cost tradeoff parameter to 10, while we keep all other hyperparameters as proposed in~\cite{berthelot2019mixmatch}. We train MixMatch for $10^5$ iterations with Adam using batch size 64 and linearly decaying learning rate from $10^{-3}$ to $10^{-5}$. 

We evaluate MixMatch by comparing to supervised training on a small number of labeled samples chosen uniformly at random, where each class is represented by at least one sample. Figures \ref{fig:fsdd-uniform}-\ref{fig:speech-commands-uniform} show $20\%$ and $35\%$ gaps between MixMatch and supervised training in the test accuracy on the two datasets. Note that training the same model on the full labeled training set achieves $100\%$ test accuracy on Free Spoken Digit Dataset and $96\%$ on Speech Commands, respectively. These large performance gaps between supervised and semi-supervised training motivate that pool-based active learning should be leveraged together with SSL.

We compare our proposed batch acquisition strategy to other batch selection methods for active learning with SSL, including $k$-center selection based on last layer embeddings~\cite{sener2018active}, consistency-based batch selection~\cite{gao2019consistency}, max-entropy selection and random sampling. For max-entropy selection, we select the top $b$ samples with highest predictive entropy, where predictions are averaged over 2 data augmentations. For the consistency-based selection~\cite{gao2019consistency} we employ 5 augmentations to calculate the variance for the predictions.  We use each acquisition strategy with the model trained with MixMatch as presented above.
In each active learning round, we retrain the model from scratch and we query labels for a batch of size $b=10$. We test the methods in the small labeled pool regime, where the start / end labeled pool size is 10 / 60 for Free Spoken Digit and 50 / 200 for Speech Commands --- we ensure  that each class is represented in the initial pool.  

For our method, we calculate the NTK corresponding to the Wide ResNet-28 (infinite-width limit) without batch normalization and pooling using the library of Novak et al.~\cite{neuraltangents2020} and set the number of Nystr\"om features to $m=2000$. In the experiments, we found it beneficial modify our strategy by selecting 90\% of the queried points by Algorithm \ref{alg:bilevel_batch_al} and selecting the remaining 10\% uniformly at random in each active learning round.

 Figures \ref{fig:fsdd}-\ref{fig:speech-commands} show the test accuracy obtained by different selection methods as a function of the number of labeled samples, while the final test accuracy is also shown in Table \ref{table:res}. We observe that, in our small labeled pool setting, the majority of the methods suffer from selecting redundant points in the batch (i.e., selecting multiple similar points with the same true class label per batch) and thus underperforming compared to uniform sampling. While the consistency-based selection performs well on the Free Spoken Digit Dataset, it suffers from lack of diversity in the queried batches on the Speech Commands and thus performs significantly worse than uniform sampling. A similar observation was made by the authors~\cite{gao2019consistency} in the small labeled pool regime in the image domain. We also experimented with starting the acquisition method only after $150$ labels have been acquired randomly, giving the better result for consistency-based selection reported in Table \ref{table:res}.

\looseness -1 The results in Figures  \ref{fig:fsdd}-\ref{fig:speech-commands} confirm that our method (referred to as ``Bilevel") is highly effective even with a few labeled samples, with a performance comparable to the next-best method's performance with $30\%$ more labels. To  illustrate the diversity of the batches queried by our method, we plot the first round acquisitions on Free Spoken Digit Dataset in Figure \ref{fig:bilevel-diversity}, in which points are visualized by t-SNE \cite{maaten2008visualizing} on the last layer embeddings of the model trained on the initial pool (55\% test accuracy). Our method selects a diverse batch representing 9 classes, where 9 out of 10 samples are misclassified by the model.
\vspace{-1mm}
\begin{table}[h]
\centering
    \caption {Semi-supervised batch active learning  with batch size of 10. Results reported at 60 acquired labels for Free Spoken Digit and 200 for Speech Commands. \label{table:res}}
\begin{tabular}{@{}ccc@{}}
\toprule
\textbf{\begin{tabular}[c]{@{}c@{}} \\ Method\end{tabular}} & \textbf{\begin{tabular}[c]{@{}c@{}}Free\\ Spoken Digit\end{tabular}} & \textbf{\begin{tabular}[c]{@{}c@{}}Speech \\ Commands\end{tabular}} \\ \midrule
Uniform                                                                       & 91.33 $\pm$ 3.98                                                     & 88.13 $\pm$ 0.79                                                    \\
Max-Entropy                                                                   & 87.56 $\pm$ 7.44                                                     & 78.52 $\pm$ 4.62                                                    \\
$k$-center~\cite{sener2018active}                            & 92.67 $\pm$ 2.37                                                     & 81.45 $\pm$ 3.20                                                    \\
Consistency~\cite{gao2019consistency}                        & 96.61 $\pm$ 2.80                                                     & 82.06 $\pm$ 3.98                                                                    \\
\textbf{Bilevel (ours)}                                                                & \textbf{98.89 $\pm$ 1.13}                                                     & \textbf{90.58 $\pm$ 0.97}                                                    \\ \bottomrule
\end{tabular}
\vspace{-4mm}
\end{table}

\section{Conclusion}
\label{sec:conclusion}
We presented a novel batch acquisition strategy for pool-based active learning when the model training is performed in a semi-supervised manner. We formalized our approach as a  cardinality-constrained bilevel optimization problem and provided a reformulation suitable for deep neural networks trained with data augmentations. We demonstrated the empirical effectiveness of our method on keyword detection tasks, where we observed significant performance gains in the regime of working with a small labeled data pool.

\section{Acknowledgements}
 This research was supported by the SNSF grant \seqsplit{407540\_167212} through the NRP 75 Big Data program and by the European Research Council (ERC) under the European Union’s Horizon 2020 research and innovation programme grant agreement No 815943.

\newpage
\bibliographystyle{abbrv}
\bibliography{refs}

\end{document}